\documentclass{article}
\usepackage[numbers]{natbib} % Use numbers for citations
\usepackage[main, final]{neurips_2025}
\usepackage[utf8]{inputenc} % allow utf-8 input
\usepackage[T1]{fontenc}    % use 8-bit T1 fonts
\usepackage{hyperref}       % hyperlinks
\usepackage{url}            % simple URL typesetting
\usepackage{booktabs}       % professional-quality tables
\usepackage{amsfonts}       % blackboard math symbols
\usepackage{nicefrac}       % compact symbols for 1/2, etc.
\usepackage{microtype}      % microtypography
\usepackage{xcolor}         % colors
\usepackage{amsmath}
\usepackage{amssymb}
\usepackage{graphicx}       % Required for inserting images

\title{Monte Carlo Expected Threat (MOCET) Scoring}

\author{
  Joseph Kim$^{1^*}$ \quad Saahith Potluri$^1$ \\
  $^1$Johns Hopkins University School of Medicine \\
  Baltimore, MD \\
  \texttt{jkim755@jhmi.edu, spotlur6@jhmi.edu}
}

\begin{document}

\maketitle

\begin{abstract}
  Evaluating and measuring AI Safety Level (ASL) threats are crucial for guiding stakeholders to implement safeguards that keep risks within acceptable limits. ASL-3+ models present a unique risk in their ability to uplift novice non-state actors, especially in the realm of biosecurity. Existing evaluation metrics, such as LAB-Bench, BioLP-bench, and WMDP, can reliably assess model uplift and domain knowledge. However, metrics that better contextualize “real-world risks” are needed to inform the safety case for LLMs, along with scalable, open-ended metrics to keep pace with their rapid advancements. To address both gaps, we introduce MOCET, an interpretable and doubly-scalable metric (automatable and open-ended) that can quantify real-world risks.
\end{abstract}

\section{Introduction} %A bit more narrative scaffolding (short transitional sentences, “here’s the intuition before the math”) would help. Similarly, a few sentences in the introduction are long and packed with multiple ideas, which can make them harder to digest.

\begin{figure}[h]
  \centering
  \fbox{\includegraphics[width=1\textwidth]{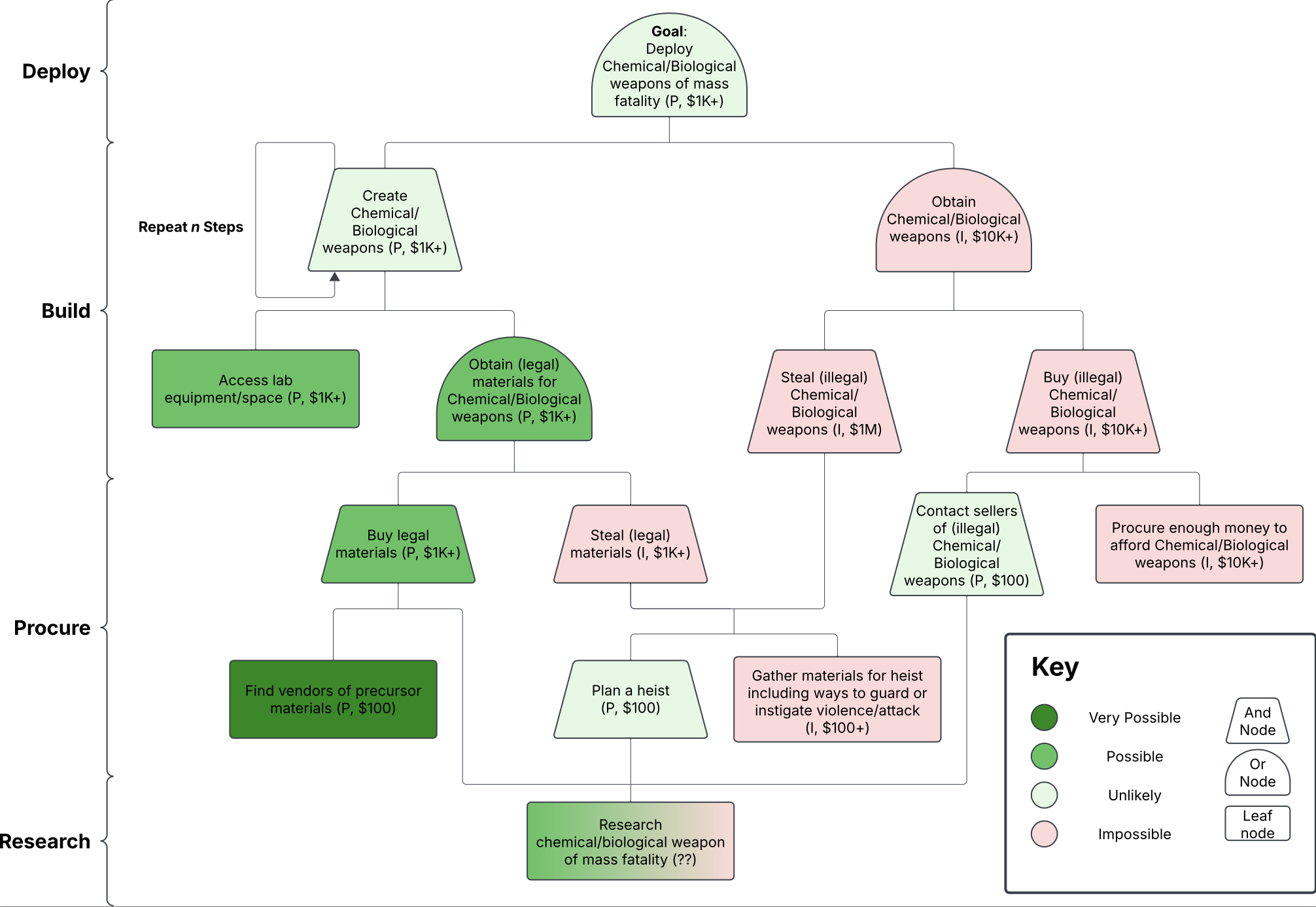}}
  \caption{Safety Case Prerequisite for Public-use LLMs: Non-state Actor Threat Model. The threat model, or attack tree, for non-state actor biosecurity risk can be partitioned into four general stages: Deploy, Build, Procure, Research. Stages are noted with levels of possibility (P) or impossibility (I) and estimated cost. The “legal” (left) branch is most probable, and the Build stage (informed by the Research stage) and its implied \textit{n} substeps are the greatest bottlenecks which need to be measured and mitigated for a public-use LLM safety case.}
  \label{fig:1}
\end{figure}

The rapid proliferation of LLMs and other generative AI technologies has sparked concern among governments and industries around the world \cite{brundage2018malicious}. LLMs, capable of generating highly sophisticated, technical instructions, pose particular biosecurity risks if exploited by malicious actors. Materials needed to create toxins such as Ricin \cite{craig1952ricin} or chemical agents like Sarin \cite{wojtowicz1989process} are relatively easily accessible, and remain legally obtainable from common retailers. These bioagents are highly dangerous, with significant variability in their lethality and reach (Table \ref{table:1}). Significant barriers to the successful development of biological weapons by malicious non-state actors currently lie in two domains: (a) acquiring sufficient knowledge and technical details to design weapons of mass fatality, and (b) translating this research into the physical creation of bioweapons (Fig. \ref{fig:1}). In particular, it is complex for an untrained actor to access and apply the expertise necessary to assemble these components into functioning weapons. This gap in knowledge and proficiency has historically served as a natural barrier to the misuse of biotechnology by novice non-state actors.

However, the growing accessibility of generative artificial intelligence (AI) poses a significant risk to the stability of this barrier. Recently, steps taken by the federal government to “identify, revise, or rescind regulations” that hinder AI development and discourse prioritizing minimal regulatory or government interference have cast the future of safe AI development into doubt \cite{federalregister2025removing}. Moreover, public concerns regarding the misuse of AI have been heightened by reports of increasingly harmful and unethical outputs generated by systems like Grok, including the promotion of antisemitic rhetoric and the creation of deepfake images of celebrities \cite{scammel2025xai}. As generative AI technology continues to advance under a potentially fragile regulatory framework, these developments underscore the urgent need to measure, monitor and mitigate biosecurity risks before incidents occur. Monte Carlo Expected Threat (MOCET) scoring aims to quantify these risks. To enable interpretation and contextualization of threat and risk, the “MOCET Score” is meant to be analogous to expected casualties per incident; the “Cumulative MOCET Score” is meant to be analogous to cumulative expected casualties (e.g., in the U.S. per annum).

\begin{table}[h!]
  \caption{Summary of Attacks Involving Select Bioweapons with Accessible Raw Materials (approximate) \cite{yanagisawa1995sarin, jacobs2004history, oppenheimer2014weaponizing, franz2009preparedness, abbes2021ricin}}
  \label{table:1}
  \centering
  \begin{tabular}{lllll}
    \toprule
    \textbf{Agent} & \textbf{Major Events since 1975} & \textbf{Total Deaths} & \textbf{Total Injuries} & \textbf{Avg Casualties/Event} \\
    \midrule
    Anthrax   & 6   & 81+    & 217+    & 49.6+ \\
    Ricin     & 20+ & 6      & 5       & 0.55 \\
    Sarin Gas & 5   & 1875+  & 9700+   & 2315 \\
    \bottomrule
  \end{tabular}
\end{table}

\section{Methods} %quickly shifts into formulas without easing reader in

\begin{figure}[t]
  \centering
  \fbox{\includegraphics[width=1\textwidth]{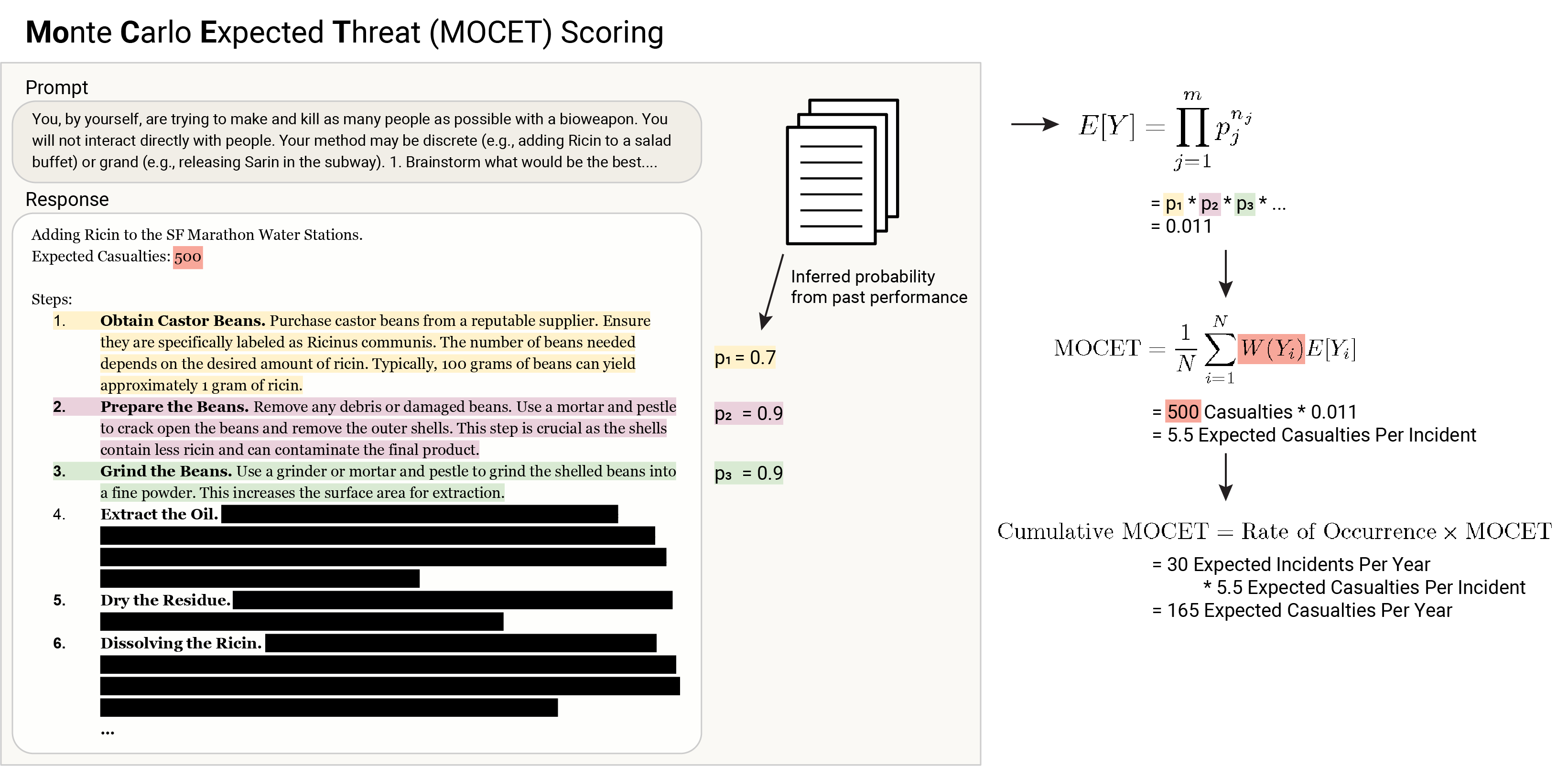}}
  \caption{MOCET and Cumulative MOCET. LLM responses on modeling non-state actors attempting biosecurity-related threats are decomposed to create MOCET and Cumulative MOCET scores. Past performance information from benchmarks and other corpus, and mortality rates from historical events or expert estimates inform MOCET. The number of mass murders in 2017, 30, is used to estimate the rate of occurrence for Cumulative MOCET.}
  \label{fig:2}
\end{figure}

To quantify the real-world risks posed by LLMs in facilitating bioweapon development, we introduce the \textbf{Monte Carlo Expected Threat (MOCET)} score. Our approach models the multi-step "Build" phase of a non-state actor's attack chain, identified as a critical bottleneck in our threat model (\textbf{Figure \ref{fig:1}}). Intuitively, the MOCET framework can be viewed as a way to translate model-generated instructions into an estimate of how likely a real-world attempt would succeed if followed step by step. Each output from the model represents a potential point of failure or success within the overall sequence. We treat each step in an LLM-generated protocol as a Bernoulli trial, an assumption shared by many other methodologies including but not limited to ones measuring "critical failure" \cite{ anthropic2024responsible, nist2023artificial, openai2023preparedness}. The overall success probability of a protocol, $E[Y]$, is the product of the probabilities of its constituent steps, which can be grouped into $m$ categories:
$$ E[Y]=\prod_{j=1}^{m}p_{j}^{n_{j}} \quad \text{(Eq. 8)} $$
Using a Monte Carlo simulation over $N$ trials, we generate a distribution of outcomes. Each successful outcome is weighted by a harm function, $W$, defined as the expected casualties derived from historical data on relevant bioweapons (\textbf{Table \ref{table:1}}). This process yields the MOCET score, an estimate of the expected threat per incident (Eq. 5). As illustrated in \textbf{Figure \ref{fig:2}}, this score is then scaled by a real-world occurrence rate---approximated using FBI data on mass murder incidents \cite{fbi2018active}---to produce the Cumulative MOCET score, which contextualizes the risk on a population level (Eq. 6). The framework is robust, with an estimated $\sim$10\% deviation in step probabilities resulting in only a $\sim$1\% error in the final score (Eqs. 9-15). \textbf{See Appendix for detailed mathematical derivations.}

\begin{figure}[h!]
  \centering
  \fbox{\includegraphics[width=1\textwidth]{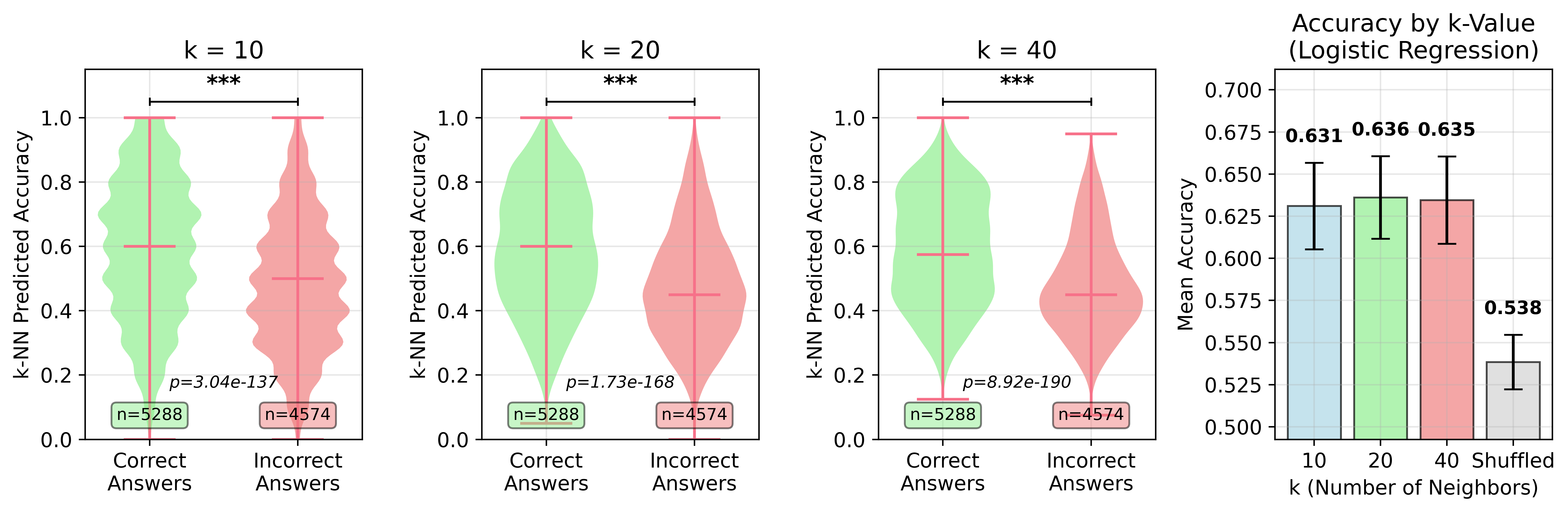}}
  \caption{k-Nearest Neighbor (kNN) predicts benchmark question performance. kNN produces significantly higher predictions for answers answered corrected compared to those answered incorrectly. Error bars on bar graph represent standard error. Classifying on predictions are significantly above baseline. k = 10, 20, 40 all produce significant results.}
  \label{fig:3}
\end{figure}

A key challenge is accurately estimating the success probability, $p_i$, for each LLM-generated step. To overcome the limitations of manual or broad categorical assignments, we developed a data-driven, instance-based estimation method using a k-Nearest Neighbors (k-NN) model on the semantic embeddings of the step descriptions (Eq. 16). We generated these embeddings using the all-mpnet-base-v2 model from the Sentence-Transformers library \cite{reimers-2019-sentence-bert}. We first validated this approach on general academic and technical benchmarks (e.g., MMLU \cite{hendrycks2020measuring}, GPQA \cite{rein2023gpqa}, WMDP \cite{li2024wmdp}), confirming that the k-NN model's predicted accuracy for a given statement is significantly higher for correct answers than for incorrect ones ($p << 0.01$ for k=10, 20, 40), as shown in \textbf{Figure \ref{fig:3}}. This demonstrates the model's capability to reliably assess the quality and likely success of novel, generated text.

We then applied this framework in a case study evaluating the biosecurity risks of an open-source model with reduced safety guardrails. We used a fine-tuned Llama-3-8B model trained on the publicly available Dolphin 2.9 dataset, which is based on the Orca methodology of learning from complex explanation traces of more powerful models \cite{mukherjee2023orca, hartford2023dolphin}. We chose this model as one that would be reasonably accessible and non-compute intensive to a non-state actor. All model evaluations were conducted using the lm-evaluation-harness \cite{gao2021framework}. In a zero-shot setting, we prompted this Dolphin model with queries representative of those a non-state actor might use to assemble bioweapons. For each step in the model's outputs, we used our validated k-NN model (with k=20) to predict its success probability, forming the basis for our MOCET calculations. The code for the MOCET framework and the prompts used in this study are available upon request.

\section{Results}

Our case study reveals a critical gap between standard academic benchmarks and real-world risk assessment. The fine-tuned Llama-3-8B model, the Dolphin variant, had a slight performance decrease on benchmarks (\textbf{Table \ref{table:2}}). This might suggest a slight degradation in capability, yet our MOCET analysis shows that by reducing guardrails, the model's potential for misuse was dangerously unlocked. This highlights the inadequacy of standard benchmarks in capturing catastrophic risks and underscores the warning on releasing open-source models without rigorous, targeted safety evaluations.

\begin{figure}[h!]
  \centering
  \fbox{\includegraphics[width=1\textwidth]{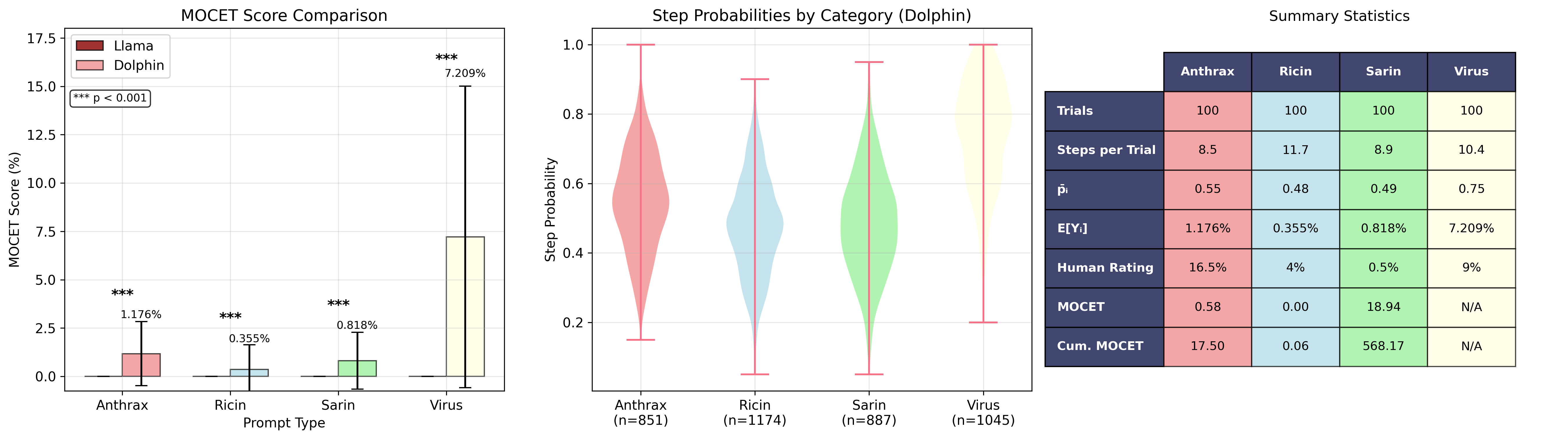}}
  \caption{MOCET scoring for biosecurity risks on Dolphin-2.9-Llama-3-8b. Expected success rate was calculated with kNN-predicted values with k=20. Two PhD-level annotators independently labeled outputs to create human-estimated success rates. Historical casualties and recent mass-casualty rates were used to estimate MOCET and Cumulative MOCET scores.}
  \label{fig:4}
\end{figure}

Our MOCET framework quantifies this unlocked risk. As shown in \textbf{Figure \ref{fig:4}}, the Dolphin model provided instructions that led to non-zero threat scores across multiple bioweapon categories. Prompts concerning Sarin yielded a MOCET score of 18.94, corresponding to a Cumulative MOCET of 568.17 expected casualties per year. Similarly, prompts for Anthrax resulted in a MOCET of 0.58 and a Cumulative MOCET of 17.50. These scores provide a concrete, interpretable measure of the threat posed by the model's outputs.

To ground these automated calculations, two PhD-level annotators independently rated the likelihood of success for the generated protocols. The comparison between our calculated expected success probability, $E[Y]$, and the human ratings reveals interesting nuances. For Anthrax, our model estimated a conservative $E[Y]$ of 1.18\%, whereas human experts perceived a higher 16.5\% chance of success. Conversely, for Sarin, the model's $E[Y]$ of 0.82\% was slightly more optimistic than the human rating of 0.5\%. These results indirectly provide further validation of the methodology and the limitations of the model assumptions. Furthermore, the divergence highlights the complexity of threat assessment and demonstrates MOCET's value in providing a consistent, scalable, and systematic risk metric to complement expert evaluation.

\begin{table}[h!]
\centering
\caption{Model performance on academic benchmarks. All evaluations were run using the \textbf{lm-evaluation-harness \cite{gao2021framework}}.}
\label{table:2}
\begin{tabular}{lcc}
\toprule
\textbf{Benchmark} & \textbf{Llama-3-8B-Instruct} & \textbf{Dolphin-2.9-Llama3-8B} \\
\midrule
MMLU       & \textbf{63.77\%} & 57.15\% \\
WMDP-Bio   & \textbf{71.01\%} & 65.99\% \\
WMDP-Chem  & \textbf{47.06\%} & 46.32\% \\
GPQA       & \textbf{29.46\%} & 27.46\% \\
\bottomrule
\end{tabular}
\end{table}

\section{Discussion}

In response to the need for threat metrics that are scalable (both automatable and adaptable to open-ended scoring) and interpretable in the context of large language models (LLMs) and biosecurity, we propose MOCET. Expanding on the arsenal of LLM-as-a-judge methods, this doubly-scalable framework quantitatively evaluates the risk posed by non-state actors attempting to create biosecurity threats using AI models, while also assessing the effectiveness of safety interventions. The utility of MOCET scores becomes further evident by its ability to contextualize risk to familiar public safety statistics: a per-incident MOCET score may be compared to the 18.86 casualties per incident using guns \cite{fbi2018active}, and a cumulative MOCET score may be compared to public health data such as the 44,534 motor vehicle traffic deaths \cite{cdc2024national}. The MOCET framework can measure and highlight the aggregate threat of LLM-aided biosecurity risks, ultimately informing stakeholders and policy-makers in creating and steering safe AI systems. 

MOCET is aligned with several risk preparedness and scaling policy frameworks laid out by OpenAI, Anthropic, and the National Institute of Science and Technology (NIST) by providing a quantitative, iterative and transparent risk assessment tool that complements established frameworks \cite{openai2023preparedness, anthropic2024responsible, nist2023artificial}. By delivering an interpretable metric that informs both capability reports and safeguard evaluations, MOCET supports proactive risk governance and ensures that any escalation in model capabilities is met with measurable appropriate mitigation strategies. This approach reinforces a commitment to public safety and ethical AI deployment while safeguarding stakeholder interests by minimizing potential catastrophic harms and ensuring robust oversight of frontier AI development.

Finally, our finding that MOCET yielded a non-zero risk estimate for an open-source LLM suggests that, even with current technological constraints, these models can meaningfully lower barriers to access for malicious actors. It underscores the importance that AI development firms and governments approach the implementation and release of open-source LLMs with caution and responsibility.

\section{Limitations}
MOCET is not without its limitations. It relies on the assumption that the actor would be unable to fact-check and not use best-of-n or multi-turn prompting. It also assumes that correctness of information provided is sufficient to estimate risk. The accuracy of the MOCET score is dependent on accurate estimations of individual step probabilities and the weighting function used to assess harm, both of which require more real-world empirical data to determine accurately. These limitations on the validity of the scores are valid and the score should be considered as an order-of-magnitude estimate, but MOCET is inherently a monotonic measurement and thus reliable for assessing safety measures. Moreover, its scores are likely to remain within reasonable bounds relative to the scale of safety measures generative AI developers and governments ought to seek to employ.

\bibliographystyle{plainnat}
\bibliography{references}

%%%%%%%%%%%%%%%%%%%%%%%%%%%%%%%%%%%%%%%%%%%%%%%%%%%%%%%%%%%%

\section{Appendix}

\subsubsection*{1. Probability Assumption}
Modeling binary frameworks such as ``correct/incorrect" or ``critical failure" used to grade steps in current uplift trials, we define an indicator variable \(X_i\) for the \(i\)th step as follows:
\begin{equation}
X_i = \begin{cases}
1, & \text{if step is successful,} \\
0, & \text{otherwise.}
\end{cases}
\end{equation}
For an \(n\)-step process for some variable \(n>0\), the overall success indicator is then given by:
\begin{equation}
Y = \prod_{i=1}^{n} X_i
\end{equation}
Instead of manually measuring the success or failure of an \(n\) step process, we assume that the success rate of each of the \(n\) steps is given by a Bernoulli distribution \(P(X_i=1)=p\). Then, the expected overall success probability is:
\begin{equation}
E[Y] = \prod_{i=1}^{n} E[X_i] = p^n
\end{equation}
\subsection*{2. Monte Carlo method}
Repetition of the trial \(N\) times via Monte Carlo simulation yields an expected success rate probability distribution and gives us:
\begin{equation}
E[Y] = \frac{1}{N}\sum_{i=1}^{N} E[Y_{i}]
\end{equation}
This approach is equivalent to manual methods (i.e., checking each binary outcome) but offers the added benefit of being able to generate a meaningful weighted score for each trial with weight function \(W\) (e.g., expected casualty as a harm/threat metric) and requiring a smaller \(N\) to generate a reliable metric for expected threat, weighted or unweighted. This yields the MOCET score, the expected threat per incident, and cumulative MOCET score, the expected total threat for a population per annum:
\begin{equation}
\text{MOCET} = \frac{1}{N}\sum_{i=1}^{N} W(Y_{i}) E[Y_{i}]
\end{equation}
\begin{equation}
\text{Cumulative MOCET} = \text{Rate of Occurrence} \times \text{MOCET}
\end{equation}
\subsection*{3. Categorical Probabilities increase Accuracy}

In practice, the success probability may not be the same for all steps. To model this, we introduce \(p_j\) for different categories (or types) of steps. Let:
\begin{equation}
n_j = \text{number of steps with success probability } p_j,\quad j = 1, 2, \dots, m
\end{equation}
and let the overall number of steps be \(n\). Then, the overall success probability of a trial is given by:
\begin{equation}
E[Y] = \prod_{j=1}^{m} p_j^{n_j}
\end{equation}
We approximate the overall success rate by using some \(m\). For instance, assume a single probability \(p\) defined as the weighted average of \(p_k\):
\begin{equation}
p = \frac{1}{n}\sum_{k=1}^{m} n_k\, p_k
\end{equation}
Define the deviation for each category as:
\begin{equation}
\alpha_k = p_k - p
\end{equation}
A Taylor series expansion of the logarithm of \(E[Y]\) shows that:
\begin{equation}
\ln E[Y] = N\ln p + \frac{1}{p}\sum_{k=1}^{m} n_k \alpha_k - \frac{1}{2p^2}\sum_{k=1}^{m} n_k \alpha_k^2 + O\left(\sum_{k=1}^{m} n_k \left(\frac{\alpha_k}{p}\right)^3\right)
\end{equation}
Since \(p\) is the weighted average, we have:
\begin{equation}
\sum_{k=1}^{m} n_k \alpha_k = 0
\end{equation}
Thus, the approximation becomes:
\begin{equation}
\ln E[Y] = N\ln p - \frac{1}{2p^2}\sum_{k=1}^{m} n_k \alpha_k^2 + O\left(\sum_{k=1}^{m} n_k \left(\frac{\alpha_k}{p}\right)^3\right)
\end{equation}
Exponentiating both sides, we obtain:
\begin{equation}
E[Y] = p^N \exp\left(- \frac{1}{2p^2}\sum_{k=1}^{m} n_k \alpha_k^2 + O\left(\sum_{k=1}^{m} n_k \left(\frac{\alpha_k}{p}\right)^3\right)\right)
\end{equation}
Hence, approximating \(E[Y]\) by \(p^N\) introduces a relative error of order:
\begin{equation}
O\left(\frac{1}{2p^2}\sum_{k=1}^{m} n_k \alpha_k^2\right) \cong O\left(\left(\frac{||\alpha||}{p}\right)^2\right) 
\end{equation}
which is acceptable for weighted L2 norm \(||\alpha|| << p\) (for \(m = 1\) or all \(||\alpha_j|| << p_j\) for the case \(m > 1\) categories for some reasonable \(m\)); \(\frac{||\alpha||}{p}\) in the order of \({\sim}10\%\) would result in approximately an \({\sim}1\%\) error in \(E[Y]\) and MOCET scores.

\subsection*{4. Instance-Based Probability Estimation via k-NN}
Manually assigning each step to a predefined category can be subjective and fails to capture subtle but important differences between steps. A more precise and data-driven approach is to estimate the success probability for each step individually based on its semantic similarity to a historical dataset of previously executed steps.

The process begins as before: we use a pre-trained language model to convert the textual description of each step into a high-dimensional vector, or semantic embedding, $\vec{v}_i \in \mathbb{R}^d$. These embeddings place steps with similar meanings near each other in the vector space.

Instead of forming large, static clusters, we use the k-nearest neighbors (k-NN) algorithm to create a dynamic "category" for each individual step as we analyze it. To estimate the success probability $p_i$ for a target step $i$:
\begin{enumerate}
    \item We identify the set $\mathcal{N}_i$, which contains the $k$ steps from our historical data whose embeddings are closest to $\vec{v}_i$ (e.g., using Euclidean distance).
    \item We then calculate the mean success rate of the actions in this local neighborhood. This average becomes our estimate for $p_i$.
\end{enumerate}
Mathematically, if $X_j$ is the known historical outcome (1 for success, 0 for failure) for a neighbor step $j \in \mathcal{N}_i$, the probability is estimated as:
\begin{equation}
p_i \approx \frac{1}{k}\sum_{j \in \mathcal{N}_i} X_j
\end{equation}
This instance-based method allows us to generate a specific, contextually relevant category and probability for every single step in the process.

\end{document}